\documentclass[sigconf, dvipsnames]{acmart}
\usepackage{xcolor}
\usepackage[shortlabels]{enumitem}

\DeclareMathOperator*{\argmin}{arg\,min}

\AtBeginDocument{%
  \providecommand\BibTeX{{%
    \normalfont B\kern-0.5em{\scshape i\kern-0.25em b}\kern-0.8em\TeX}}}

\setcopyright{acmcopyright}
\copyrightyear{2020}
\acmYear{2020}
\acmDOI{0/0}

\acmConference[DeepSpatial'20]{San Diego '20: 1st ACM SIGKDD Workshop on Deep Learning for Spatiotemporal Data, Applications, and Systems (DeepSpatial 2020)}{August 24, 2020}{San Diego, CA}
\acmBooktitle{DeepSpatial'20: 1st ACM SIGKDD Workshop on Deep Learning for Spatiotemporal Data, Applications, and Systems (DeepSpatial 2020), August 24, 2020, San Diego, CA}
\acmPrice{15.00}
\acmISBN{978-1-4503-XXXX-X/18/06}

\begin{document}

\title{Towards Spatial Variability Aware Deep Neural Networks (SVANN): A Summary of Results}

\author{Jayant Gupta}
\email{gupta423@umn.edu}
\affiliation{%
  \institution{University of Minnesota}
   \city{Twin Cities, USA}
}

\author{Yiqun Xie}
\email{xiexx347@umn.edu}
\affiliation{%
  \institution{University of Minnesota}
   \city{Twin Cities, USA}
}

\author{Shashi Shekhar}
\email{shekhar@umn.edu}
\affiliation{%
  \institution{University of Minnesota}
   \city{Twin Cities, USA}
}

\renewcommand{\shortauthors}{J. Gupta, Y. Xie, S. Shekhar}

\begin{abstract}
Spatial variability has been observed in many geo-phenomena including climatic zones, USDA plant hardiness zones, and terrestrial habitat types (e.g., forest, grasslands, wetlands, and deserts). However, current deep learning methods follow a spatial-one-size-fits-all (OSFA) approach to train single deep neural network models that do not account for spatial variability. In this work, we propose and investigate a spatial-variability aware deep neural network (SVANN) approach, where distinct deep neural network models are built for each geographic area. We evaluate this approach using aerial imagery from two geographic areas for the task of mapping urban gardens. The experimental results show that SVANN provides better performance than OSFA in terms of precision, recall, and F1-score to identify urban gardens. 
\end{abstract}


\maketitle

\section{Introduction}
\label{sec:introduction}
Deep learning techniques have resulted in significant accuracy improvements in image based object recognition tasks \cite{krizhevsky2012imagenet,szegedy2015going}. They use multiple layers that allow approximate modeling of all continuous functions \cite{cybenko1989approximations}. Unlike traditional machine learning, which requires manual feature engineering, deep learning models interpret the data and automatically generate features \cite{erickson2017machine}. The current deep learning literature \cite{zhu2017deep,miotto2018deep,guo2016deep} follows a spatial one size fits all approach in which deep neural networks are trained without consideration of spatial variability.

\begin{figure}[htp]
  \centering
    \includegraphics[width=0.99\linewidth]{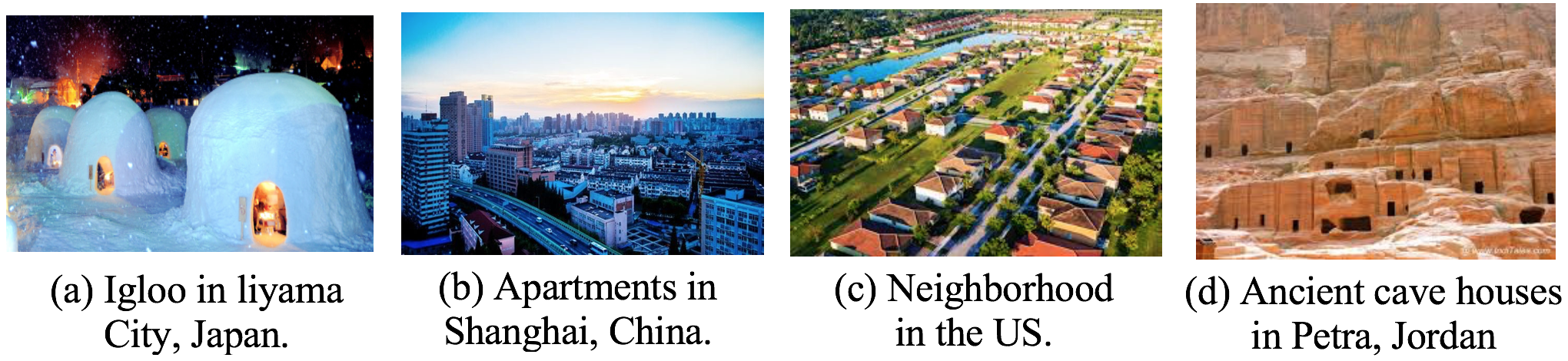}
  \caption{Spatial variability in houses and background.}
  \label{fig:spatial_variability}
\end{figure}

Geographic properties differ across different areas giving rise to varied geophysical and cultural phenomena. This spatial variability results in the lack of consistent object detection models across geographic areas. Knowledge of spatial variability is necessary to understand the spatial patterns of events and objects over an area that vary spatially \cite{turner2005causes}. The models can be affected by two types of spatial variability: the variability in the object of interest itself, which may differ in shape, size, or both; and the variability in the background of the object of interest. For example, a computational model that is trained to find residential housing in the the US may have difficulty finding houses in other places where housing construction is adapted to different local climates or other conditions (e.g., cave houses in Petra, igloos in polar regions, etc.). Here the neighbor surroundings differ as well. Figure \ref{fig:spatial_variability} shows the spatial variability in houses and their background across the globe.

Spatial variability has been observed in many geo-phenomena including climatic zones, USDA plant hardiness zones \cite{USDA_hardiness}, and terrestrial habitat types (e.g., forest, grasslands, wetlands, and deserts). The difference in climatic zones affect the plant and animal life of the region. Similarly, knowledge of plant hardiness zones helps gardeners and growers to assess appropriate plants for a region. Further laws, policies and culture differ across countries and even states within some countries. Spatial variability is considered as the second law of geography \cite{leitner2018laws} and has been adopted in regression models (e.g., Geographically Weighted Regression \cite{fotheringham2003geographically}) to quantify spatial variability, the relationship among variables across study area. In this work, we assess the effect of spatial variability on object detection models built using deep learning techniques.

Specifically, we investigate a spatial variability aware deep neural network (SVANN) approach where distinct deep neural network models are built for separate geographic areas. The paper describes alternative ways to model spatial variability, including zones and distance-weighting. It also provides descriptions of alternative ways for training and make predictions using SVANN (Section 3.2). The proposed SVANN approach was evaluated experimentally as well as via a case study for detecting urban gardens in geographically diverse high-resolution aerial imagery. The experimental results show that SVANN provides better performance in terms of precision, recall, and F1-score to identify urban gardens.

Application domains and example use cases where spatial variability is relevant and needs to be considered include wetland mapping, cancer detection, and many others. A few examples are listed in Table \ref{Table:SV_Domain}.

\begin{table}[htp]
\centering
\caption{Application domain and use-case of spatial variability.}
\label{Table:SV_Domain}
\begin{tabular}{|p{1.8cm}|p{5.8cm}|}
 \hline
 \textbf{Application Domain} & \textbf{Example use cases} \\ \hline
 Wetland Mapping & Wetlands in Florida (e.g., mangrove forest) are different from those in Minnesota (e.g., marsh)\\ \hline
 Cancer cell identification & Cancer in pathology tissue samples is known to be spatially heterogeneous \cite{heindl2015mapping} \\ \hline
 Vehicle detection & vehicle types differ across India (e.g., auto-rickshaw) and USA (SUVs) \\ \hline
 Residence detection & House types and design (e.g., igloos, huts, flat-roof, ...) differs across geographic areas \\ \hline
 Urban Agriculture & Detection of urban gardens:Urban gardens designs may vary across rural (large ones), suburban (small backyard gardens) and urban (e.g., container gardens, community gardens) areas due to differing space availability and risks (e.g., deer, rabbit, ...) \\ \hline
\end{tabular}
\end{table}

\textbf{Contributions:} 
\begin{enumerate}
    \item We propose a spatial variability aware deep neural network (SVANN) approach and illustrate various training and prediction procedures.
    \item We use SVANN to evaluate the effect of spatial variability on deep learning models during the learning processe.
\end{enumerate}
\textbf{Scope:} This paper focuses on geographic and other low-dimensional space. Generalization of the proposed approaches to model variability in high dimensional spaces is outside the scope of this paper. We use a convolutional neural network (CNN) for the experimental evaluation and case studies. Evaluation of SVANN with other types of neural networks is outside the scope of this paper.
\\\textbf{Organization:} The paper is organized as follows: Section \ref{section:approach} describes the details of SVANN along with different training and prediction procedures. Section \ref{section:evaluation} describes the evaluation framework giving details on the evaluation task, evaluation metric, dataset, and experiment design. In Section \ref{section:RD}, we present the results and a discussion of the effects of spatial variability. Section \ref{section:Related-Work} briefly discusses the relevant related work. Finally, Section \ref{section:CFW} concludes with future directions. In Appendix \ref{Appendix:Aerial}, we compare and contrast different types of aerial imagery. Then, we give details of object detection using YOLO framework in Appendix \ref{Appendix:YOLO}. We also give details on dataset development in Appendix \ref{Appendix:data_development}.

\section{Approach}
\label{section:approach}
\begin{figure}[htp]
  \centering
    \includegraphics[width=0.99\linewidth]{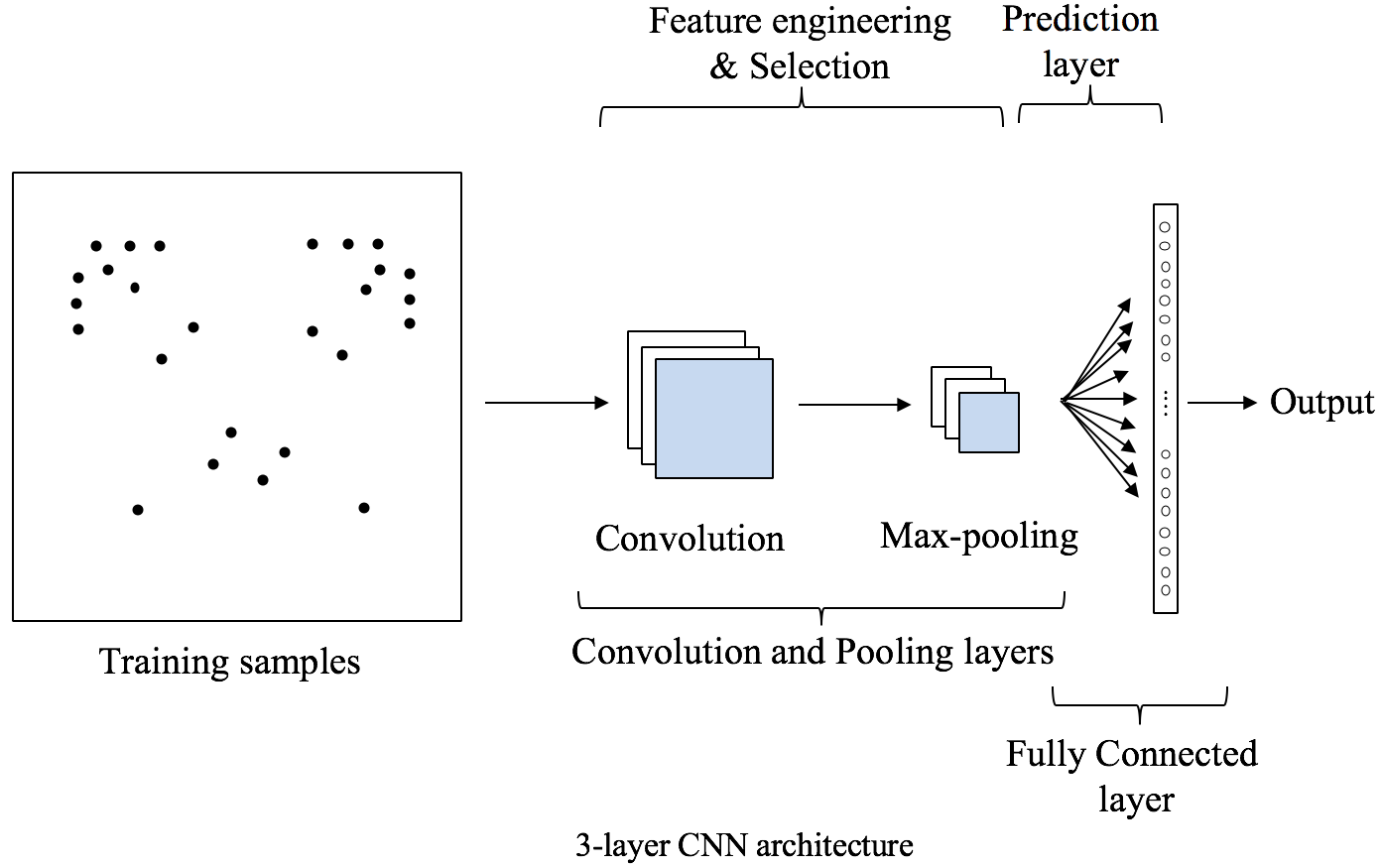}
  \caption{Spatial One Size Fits All (OSFA) approach using a CNN with 3-layers: convolution, spatial pooling, and a fully connected layer.}
  \label{fig:OSFA}
\end{figure}

In this section, we provide the details of our SVANN approach and differentiate it from the spatial one size fits all (OSFA) approach. Figure \ref{fig:OSFA} shows spatial One Size Fits All (OSFA) approach using a CNN with 3 layers: a convolution layer, a spatial pooling layer, and a fully connected layer. The initial 2 layers perform feature engineering and selection, whereas, the fully connected layer is responsible for output prediction. As shown, the approach does not account for the geographic location of training samples.
\subsection{SVANN}
\label{subsection:SVANN}
\begin{figure}[htp]
    \centering
        \includegraphics[width=0.99\linewidth]{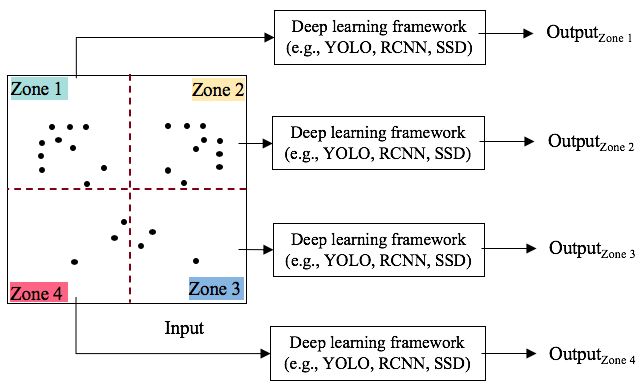}
        \caption{SVANN using fixed-partition based neighbors. Four distinct models are trained using training samples from each zone.}
    \label{fig:SVANN}
    \end{figure}

SVANN is a spatially explicit model where each neural network parameter (e.g., weight) is a function of model location $loc$. The model $f$ is composed of a sequence of $K$ weight functions or layers $(w^1(loc),...,w^K(loc))$ mapping a geographic location based training sample $x(loc)$ to a geographic location dependent output $y(loc)$ as follows, 
\begin{align}
\label{eq4}
y(loc) &= f(x(loc) ; w^{1}(loc),...,w^{K}(loc)), 
\end{align}
where $w^i(loc)$ is the weight vector for the $i_{th}$ layer. Figure \ref{fig:SVANN} shows the SVANN approach where the geographical space has 4 zones and deep learning models are trained for each zone separately. For prediction, each zonal model predicts the test sample in its zone. 

SVANN can be further classified by the choice of training and prediction procedures. Here, we describe some of these procedures.

\subsubsection{\textbf{Training:}}
\label{subsec:training}
There are at least two possible training procedures, namely, model-location dependent sampling for learning and distance weighted model-location dependent sampling for learning.

\begin{figure}[htp]
  \centering
   \includegraphics[width=0.99\linewidth]{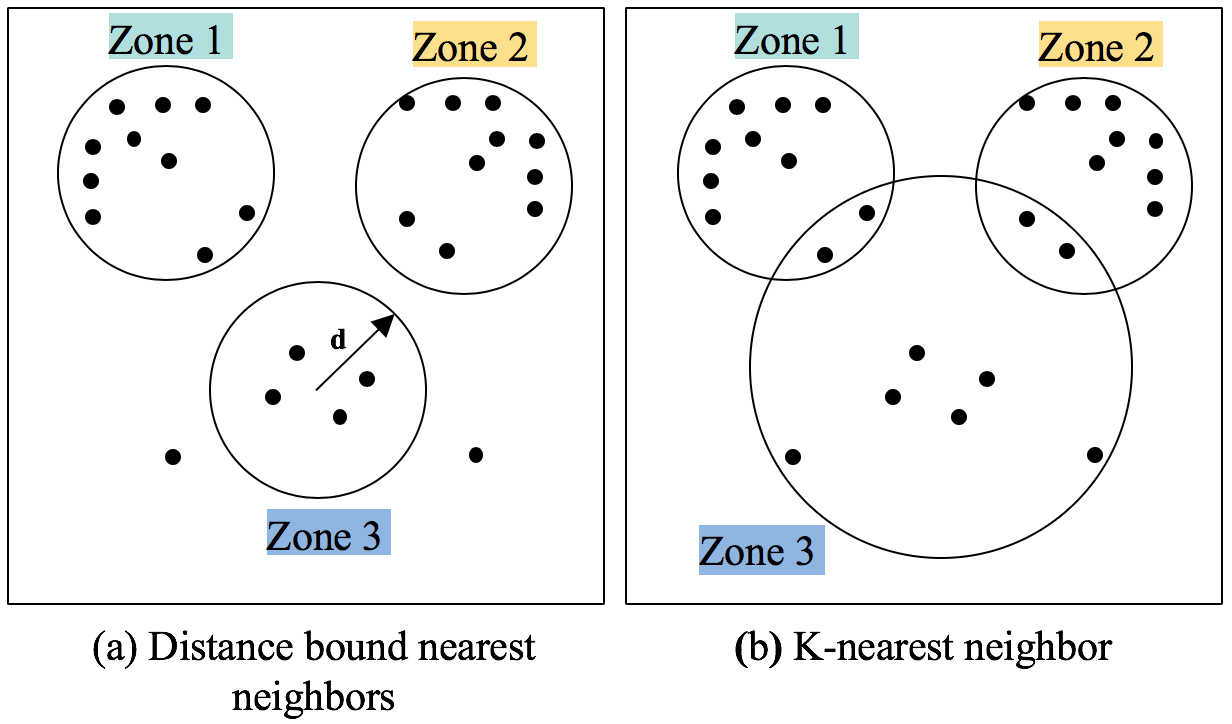}
  \caption{Model-location dependent sampling for learning.}
 \label{fig:Training-1}
\end{figure}

\textbf{1. Model-location dependent sampling for learning:} Model parameters for a location are derived by training the model using labeled samples from nearby locations. There are three types of nearest neighbor techniques that can be considered:
\begin{enumerate}[(a)]
    \item Fixed partition based neighbors: Partitions (also known as zones) are used when policies and laws vary by jurisdictions such as countries, US states, counties, cities, climatic zones. We use administrative, zonal partitions of geographical space to build individual models. This approach is simple but relatively rigid as partitions are usually disjoint and seldom change. Figure \ref{fig:SVANN} shows training SVANN models using zone based neighbors, where a sample from each zone is used to train a model for that particular zone. Partitioning the data based on zones can break up natural partitions (e.g., Zone-3 and Zone-4 in Figure \ref{fig:SVANN}.). 
   
    \item Distance bound nearest neighbors: In this training regime, a model at location ($loc_M$) is trained using nearby training samples within distance $d$. This model assumes that there are sufficient training samples in the vicinity of model locations. This approach maybe more flexible than fixed partition based approach as the training samples can overlap across models and the model locations can adapt to the spatial distribution (e.g., hotspots) of learning samples. Figure \ref{fig:Training-1}(a) shows training of different models using training samples within distance $d$.
    \item K-nearest neighbors: In this training regime, a model at location ($loc_M$) is trained using k-nearest training samples in the geographic space. This model does not assume that there are sufficient training samples in the geographic vicinity of model locations. Thus, this approach may be more flexible than distance bound nearest neighbors. Figure \ref{fig:Training-1}(b) shows training of different models using k-nearest training samples. 
\end{enumerate}
In  Model-location dependent sampling for learning (1a-1c), selected learning samples are treated equally in the training phase.
    
\textbf{2. Distance weighted model-location dependent sampling for learning:} In this training approach, all learning samples can be used to train models at different locations. To address spatial variability, nearby samples are considered more important than further away samples by adapting the learning rate. To update the neural network weights, the learning rate is multiplied by a back propagation error and a function of the distance between the selected learning sample and the location of the model. Equivalently, the learning rate depends on the distance between the labeled sample and the location for which the model is being trained. The distance function can be thought of as the inverse of distance squared as follows,
\begin{align}
    w^i(loc_M) = w^i(loc_M) + \frac{\eta}{d^2(loc_M, loc_S)}*x^i(loc_S)*\Delta y^i(loc_S)
    \label{eqn5}
\end{align}
where, $\eta$ is the learning rate, $d$ is the distance between the location of learning sample ($loc_S$) and the location of model ($loc_M$), $x^i(loc_S)$ is the input to the $i^{th}$ layer, and $\Delta y^i(loc_S)$ is the backpropagated error at layer $i$. This approach is similar to boosting techniques \cite{freund1996experiments} where weak learners or hypotheses are assigned weights based on their accuracy. It is also similar to geographically weighted regression (GWR) \cite{fotheringham2003geographically} where regression coefficients and error are location dependent. 

In the context of object detection in imagery via CNN, we note that CNN may favor nearby pixels over distant pixels (by using convolutional and pooling layers) within a single labeled sample (e.g., an 512x512 image), whereas the proposed method further favors nearby labeled samples over distant labeled samples.

\subsubsection{\textbf{Prediction:}} Since multiple models are trained at different locations and a new sample may not be at those locations we discuss two prediction methods (i.e., zonal and distance weighted voting) to combine the predictions from multiple models for the new sample.

\begin{figure}[htp]
  \centering
    \includegraphics[width=0.85\linewidth]{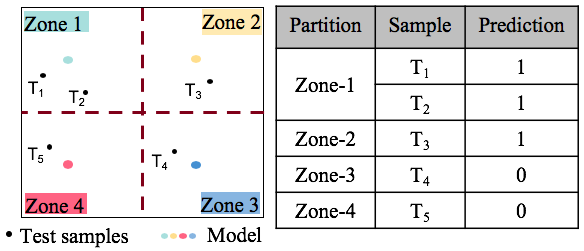}
  \caption{Zonal prediction using 5 test samples and 4 partitions.}
  \label{fig:zonal_prediction1}
\end{figure}

\textbf{1. Zonal:} Given a fixed partitioning of the geographic space (e.g., counties) prediction results from the model within the same partition will be used for prediction. If, there are multiple models within a partition, voting (e.g., majority, mean) can be used for prediction. Here the votes from all models within the partition are treated equally. Also, samples located at zone boundaries are disjoint and are assigned to a single zone. Zonal prediction is suitable for models trained on model-location dependent learning samples. Figure \ref{fig:zonal_prediction1} shows an example with 5 test samples ($T_1$ - $T_5$) and 4 partitions, where each model in a partition is a binary classifier representing classes as ($0$, $1$). The Zone 1 model is used to make prediction for test sample T1 and T2. The Zone 2 model is use to make prediction for T3 and so on. 


\begin{figure}[htp]
  \centering
    \includegraphics[width=\linewidth]{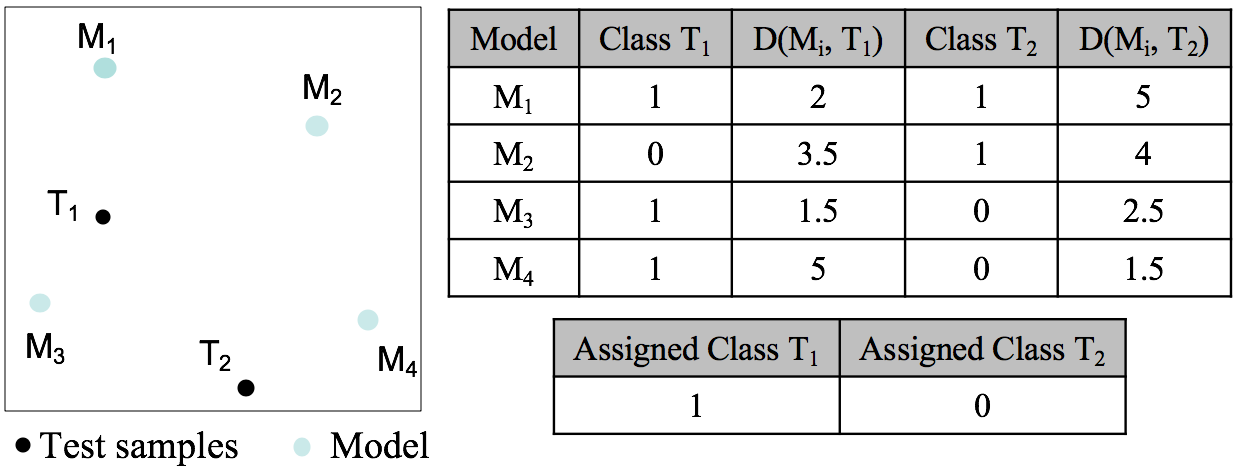}
  \caption{Distance weighted prediction using 2 test samples and 4 models. For illustration, approximate distances are used.}
  \label{fig:DW_prediction}
\end{figure}

\textbf{2. Distance weighted prediction:} Given a test sample and distances from all models, we weigh the predictions from each model as an inverse function of the distance. The highest weighted prediction is assigned as the class of the test sample. Distance weighted prediction is suitable for models trained using distance weighted model-location dependent learning samples. Figure \ref{fig:DW_prediction} shows an example with 2 test samples and 4 models where each model predicts sample class ($0$ or $1$). Assume that the adjacent (top right) table shows the predictions and distance ($D(M_i, T_i))$) of each model from test samples that are used to calculate class weights and assign class. All models are used to make a prediction for each test sample. For $T_1$, the nearest models ($M_1, M_3$) predict its class as $1$, whereas for $T_2$, the nearest models ($M_3, M_4$) predict its class as $0$ which results in final the assigned classes (shown in botton right table) for the two test samples of $1$ and $0$ respectively.
%

\subsection{Discussion}
\textbf{One Size Fit All model vs SVANN:} Given adequate learning samples and computational resources, SVANN can provide better accuracy over spatial one-size-fits-all models. Indeed, extreme cases of training a singular model may exhibit Simpson’s paradox \cite{wagner1982simpson}, where global behavior may differ from local behavior. 
\\\textbf{SVANN and number of learning samples:} SVANN need more learning samples for training to capture location specific features. However, spatial big data \cite{shekhar2012spatial} provides a wealth of data with opportunities to develop SVANN. Furthermore, citizen science \cite{silvertown2009new} provides ways where broader participation from scientists and volunteers can help generate relevant training data.
\\\textbf{Computational challenges:} In SVANN, the number of weights depends on the size of the network, number of locations, and the number of samples. This adds to the existing high computational cost of deep learning frameworks.
\\\textbf{Parametric vs Nonparametric:} A learning model that summarizes data with a set of parameters of fixed size (independent of the number of training examples) is called a parametric model. In contrast, the number of parameters in non-parametric models is dependent on the dataset \cite{russell2002artificial}. In general, SVANN can be a non-parametric model if the number of locations is not constrained. However, in special cases locations may be constrained to a fixed number of zones (e.g., US states, countries) to create parametric SVANN models.
\\\textbf{Using SVANN to assess spatial variability in a phenomenon:} If OSFA and SVANN have similar performance on a task then, it will not support the existence of spatial variability in a phenomenon. However, if SVANN outperforms OSFA then the results support spatial variability hypothesis in the phenomenon.
\\\textbf{Spatial partitioning:} The proposed training procedures do not need partitioning of input training samples. In Fixed partition based neighbors training approach (Section \ref{subsec:training} 1a.), partitions are given as input or are part of application domain. For example, COVID-19 models are built based on political boundaries (e.g., countries). In other situations, application domain may be willing to explore data-driven (e.g., spatial characteristics) partitioning, or may depend on the underlying task, which can be explored in future work. In addition, hierarchy may add further benefits and is relevant only when partitioning is needed.

\section{Evaluation Framework}
\label{section:evaluation}
This section details the evaluation framework for the SVANN approach. We explain the evaluation task and metric. We then describe the dataset categorized by object characteristics, including conversion from satellite imagery and manual annotation. Finally, we describe the experiment design including the computing resources used for experiments. Further details are given in the Appendix.

\subsection{Evaluation Task Definition}
\label{section:UGD}
\begin{figure}[htp]
  \centering
    \includegraphics[width=0.99\linewidth]{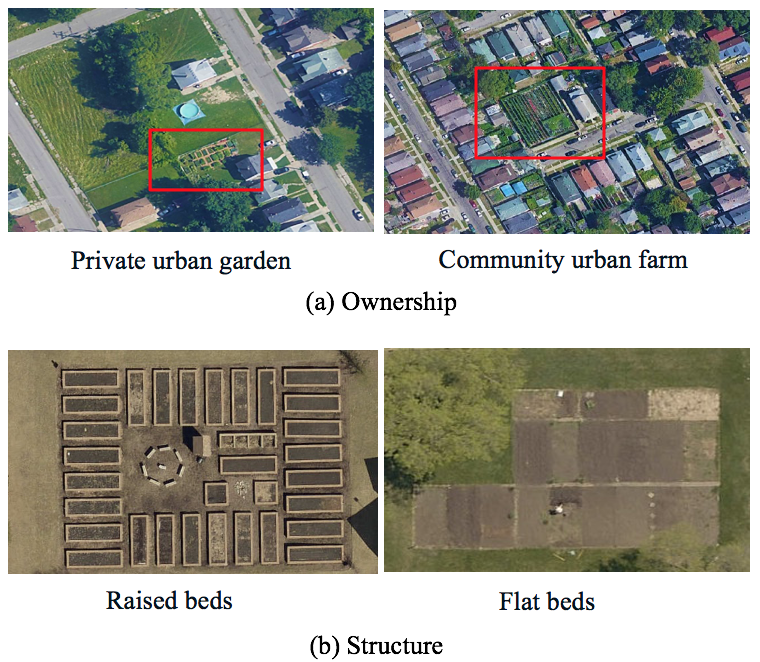}
    \caption{Type of urban gardens. (Best in color)}
  \label{fig:UGT}
\end{figure}

An urban garden is defined as a specific piece of land that is used to grow fruits or vegetables. Area-based knowledge of urban gardens aids the development of urban food policies, which currently place a strong focus on local food production. Urban gardens can be divided into many types based on ownership and structure. Figure \ref{fig:UGT}(a) shows gardens based on two types of ownership, private backyard urban garden and community urban farm. Structurally, urban gardens fall into three categories, raised beds, open fields, and rooftop gardens \cite{brown2017remote}. Figure \ref{fig:UGT}(b) shows two types of gardens based on the physical structure of their beds. As seen, raised beds are highly distinctive due to chracteristics such as surrounding stone walls. In open fields, the distinction in boundaries is relatively low, resulting in lower visual variation of garden and the surrounding area.

\begin{figure}[htp]
  \centering
     \includegraphics[width=\linewidth]{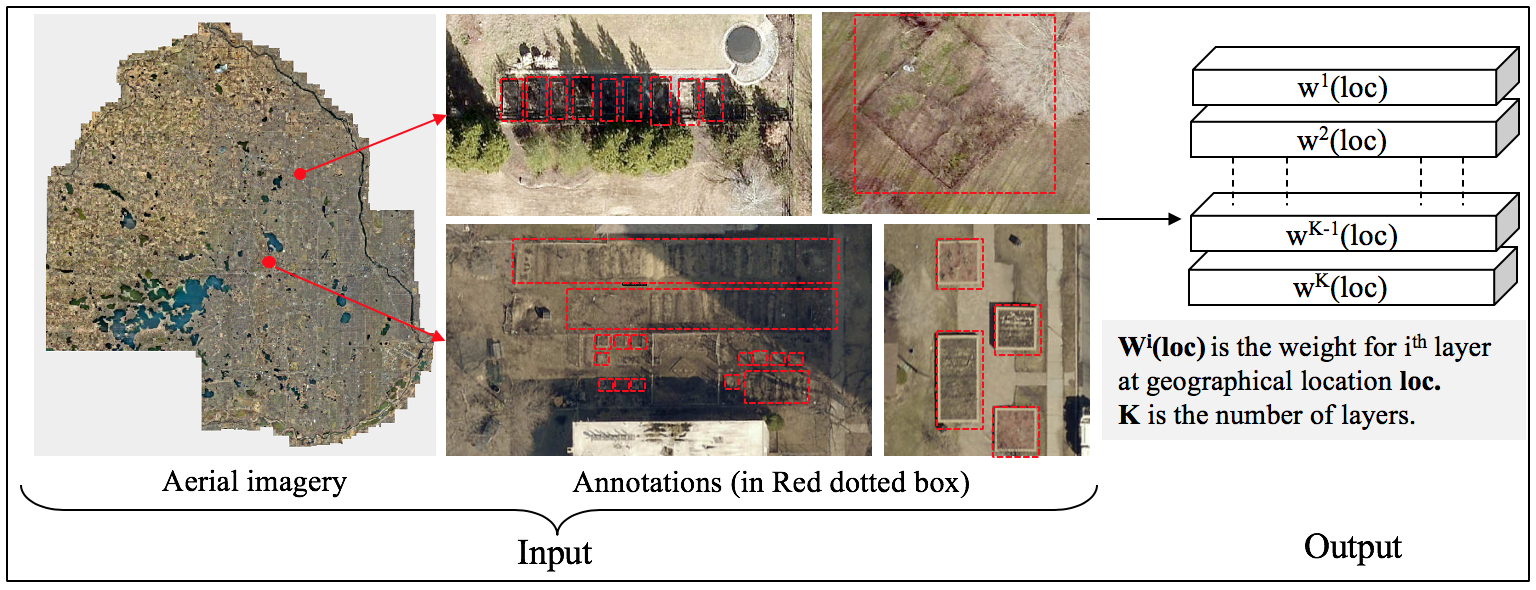}
  \caption{Example Input and output for the SVANN Evaluation Task. (Best in color)}
  \label{fig:IO}
\end{figure}

Given aerial images from different places and an object definition (e.g., urban garden), we build a computational model to detect the object having high precision and recall. There are four key constraints to the task. First, spatial variability can make a spatial one size fits all approach unusable and may require training different models at multiple locations. Second, the imagery encompasses a large geographical area (order of 1000 km$^2$) that is hard to observe manually and computationally. Third, the object in this work has no specific features and is only defined by its function to the application domain. Thus, to find and mark these objects for training is hard and can result in ambiguous annotations. Fourth, the objects of interest are in close proximity to taller neighboring objects (e.g., buildings, houses, etc). These neighboring objects cast their shadow depending on the direction of the sun, which results in partial or complete occlusion of the objects. Figure~\ref{fig:IO} illustrates the task, where the red dotted boxes are the annotations.

\subsection{Evaluation metric}
\label{subsection:Metric}
\begin{figure}[htp]
  \centering
   \includegraphics[width=\linewidth]{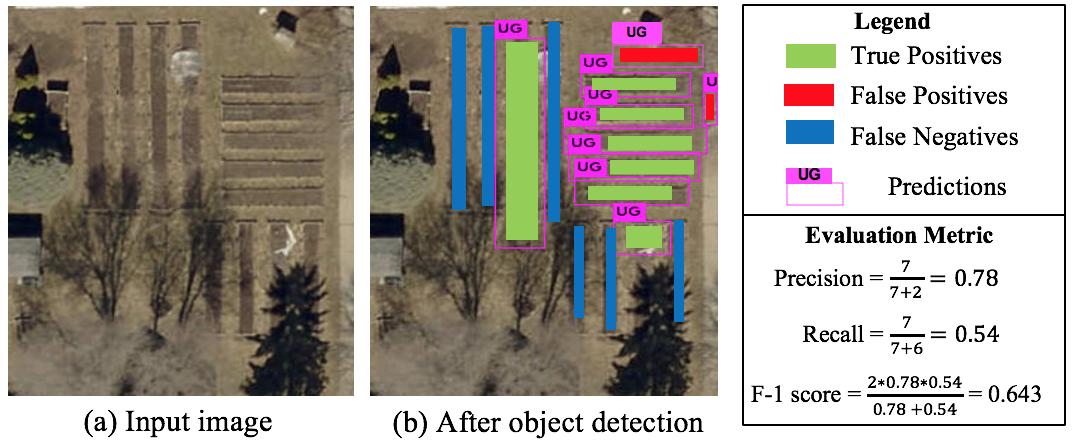}
   \caption{Illustration of true positives, false positives, and false negatives (Best in color).}
  \label{fig:TPFP}
\end{figure}

We use the F-1 metric \cite{manning2008introduction} to evaluate the results. The metric is defined as a function of precision and recall where precision is the ratio of true objects detected to the total number of objects predicted by the classifier, and recall is the ratio of true objects detected to the total number of objects in the data set. Precision and recall can be written as the function of True Positives (TP), False Positives (FP), and False Negatives (FN). Figure \ref{fig:TPFP} illustrates TP, FP, and FN in image based detection results, where Figure \ref{fig:TPFP}(a) is the input image and Figure \ref{fig:TPFP}(b) is the detection result. The detection results are color coded (as shown in the legend) to mark the TP, FP, and FN that are used to calculate the precision, recall, and F-1 score. True negatives i.e. objects other than urban gardens (e.g., Roads, Houses, Trees) were not defined. Hence, metrics using true negatives (e.g., accuracy) were not used for evaluating the results.

\subsection{Dataset}
\label{subsection:dataset_development}
\begin{figure}[htp]
  \centering
   \includegraphics[width=0.6\linewidth]{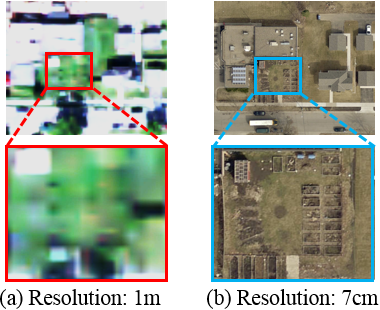}
   \caption{Urban garden in 1m and 7cm resolution imagery. (Best in color) \cite{xie2018transforming}.}
  \label{fig:High_res}
\end{figure}

The initial dataset \cite{Boyer2020} had object samples from the Minneapolis and St. Paul, MN region that were created using 2015 Google Earth imagery. Due to lower visual variation of garden and the surrounding area higher resolution imagery is preferred (Figure \ref{fig:High_res}). Hence, the samples were converted to high spatial resolution using high resolution aerial imagery \cite{DATA_HC}. The details on conversion can be found in Appendix \ref{Appendix:Conversion}. Our team also travelled to hundreds of urban gardens in Minneapolis for ground truth verification. 

\begin{figure}[htp]
  \centering
   \includegraphics[width=0.8\linewidth]{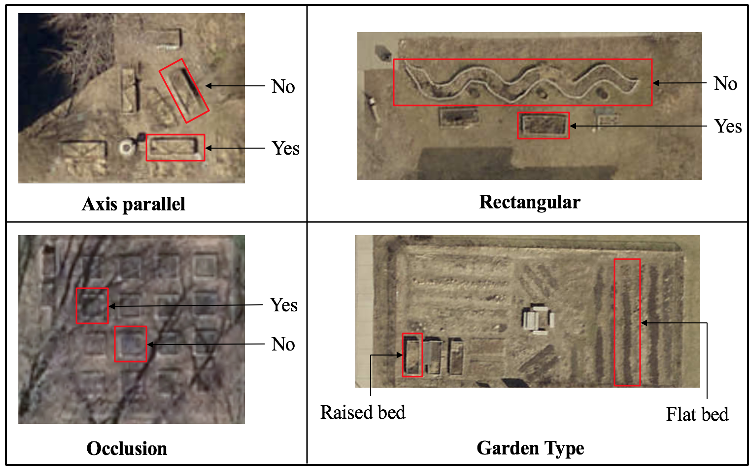}
   \caption{Object characteristics.}
  \label{fig:OC}
\end{figure}

We used ArcGIS to browse the Fulton county high resolution imagery \cite{DATA_FC} and annotate the objects. The annotated aerial imagery was then used to train the models. The details on annotation sequence can be found in Appendix \ref{Appendix:Annotation}.
\\\textbf{Object characteristics:} Besides annotations, the following generic and application specific characteristics were used for finer analysis of the results (Figure \ref{fig:OC}).\\
-- \textbf{Axis parallel (Yes/No):} Axis parallel objects allow rectangular annotations with reduced background (i.e., false positives) and have better recognition ability. This feature was recorded for every object to assess the model's efficiency on non-axis parallel objects.\\
-- \textbf{Rectangular (Yes/No):} Objects that have a clear rectangular shape can be distinctly observed as man-made, which allows better recognition. However, urban gardens can have distinctive shapes that can be geometrical or have curvy boundaries. Hence, we marked the gardens that were (approximately) rectangular from those that were clearly non-rectangular.\\
-- \textbf{Occlusion (Yes/No):} Objects that have restricted view due to neighboring buildings and trees are harder to detect and may result in lower recall values. Thus, to assess the recall values occlusion was recorded.\\
-- \textbf{Object Type (Flat/Raised):} Urban gardens can have raised beds or flat beds. Raised beds are usually accompanied by distinctive stone boundaries that improve the recognition. We marked the garden type to analyse the model performance based on their type.

For this work we annotated $1314$ urban gardens from Hennepin County, Minnesota, US and $419$ urban gardens from Fulton County, Georgia, US. The annotated images were divided into train ($80\%$) and test ($20\%$) sets for training and testing the assessment of spatial variability. The data was categorized based on the object characteristics. Table \ref{Table:HC_data} and Table \ref{Table:FC_data} show the dataset details for the two regions. For reproducibility, the dataset used in this work and relevant code are provided \cite{dataset_code}.

\begin{table}[htp]
\small
\centering
\caption{Train and test data for Hennepin County, MN categorized by object characteristics.}
\label{Table:HC_data}
\begin{tabular}{|p{1cm} p{1cm} p{1cm} p{1cm} p{1cm}| }
 \hline
 & \multicolumn{2}{c}{\textbf{Axis parallel}} & \multicolumn{2}{c|}{\textbf{Rectangular}} \\ \hline
  & Yes & No & Yes & No \\ \hline
 Train & 611 & 217 & 761 & 67  \\
 Test & 381 & 105 & 450 & 36 \\ \hline
 Total & 992 & 322 & 1211 & 103 \\ \hline
 & \multicolumn{2}{c}{\textbf{Occlusion}} & \multicolumn{2}{c|}{\textbf{Garden Type}}\\ \hline
 & Yes & No & Flat & Raised \\ \hline
 Train & 170 & 658 & 353 & 475\\
 Test & 116 & 370 & 213 & 273 \\ \hline
 Total & 286 & 1028 & 566 & 748 \\ \hline
\end{tabular}
\end{table}
\begin{table}[h!]
\small
\centering
\caption{Train and test data for Fulton County, GA categorized by object characteristics.}
\label{Table:FC_data}
\begin{tabular}{|p{1cm} p{1cm} p{1cm} p{1cm} p{1cm}| }
 \hline
 & \multicolumn{2}{c}{\textbf{Axis parallel}} & \multicolumn{2}{c|}{\textbf{Rectangular}} \\ \hline
  & Yes & No & Yes & No \\ \hline
 Train & 89 & 199 & 279 & 9 \\
 Test & 77 & 54 & 125 & 6  \\ \hline
 Total & 166 & 253 & 404 & 15 \\ \hline
 & \multicolumn{2}{c}{\textbf{Occlusion}} & \multicolumn{2}{c|}{\textbf{Garden Type}} \\ \hline
 & Yes & No & Flat & Raised \\ \hline
 Train & 88 & 200 & 12 & 276 \\ \hline
 Test & 27 & 104 & 9 & 122 \\ \hline
Total & 115 & 304 & 21 & 398 \\ \hline
\end{tabular}
\end{table}

To train well, CNNs require a large number of training data. However, creation of labeled data is expensive, leading to the use of transfer learning and data augmentation. We used Microsoft COCO~\cite{lin2014microsoft} for transfer learning. It is an extensive dataset that has around $200K$ labeled images with around $1.5$ million object instances divided into $80$ object categories. Further, the framework used to evaluate SVANN in this work uses random crops, color shifting, etc for data augmentation \cite{redmon2017yolo9000}.

 %
\subsection{Experiment design}
\label{subsection:ED}
\begin{figure}[htp]
  \centering
   \includegraphics[width=\linewidth]{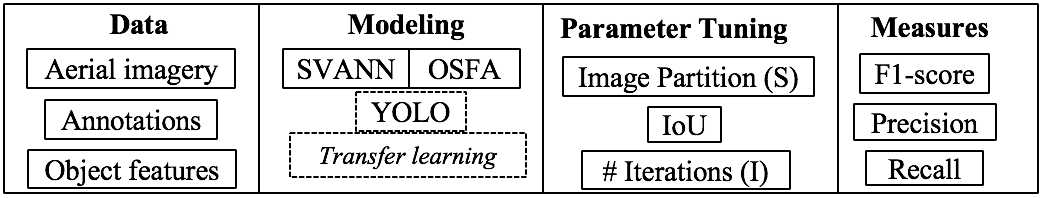}
   \caption{Experiment design}
  \label{fig:Experiment_Design}
\end{figure}

Since our goal here is a proof of concept, we limited our experiments to the special case of two types of training approaches. We trained individual models for two disjoint and distant geographic regions (i.e., Hennepin county, MN and Fulton County, GA). Due to rigid boundaries it is the base case of fixed partition based neighbors where number of partitions is 2. Furthermore, due to the large distance between the two regions, the samples are not neighbors. Overall, we trained and compared three models where, Model-1 (Hennepin County, MN) and Model-2 (Fulton County, GA) were trained separately on imagery data from different geographical areas; Model-3 was based on a spatial One Size Fits All (OSFA) approach that was trained on imagery data from both areas together. Figure~\ref{fig:Experiment_Design} shows the experiment design composed of 4 key parts, Data, Modeling, Parameter tuning, and model evaluation measures. 

\begin{figure}[htp]
  \centering
   \includegraphics[width=\linewidth]{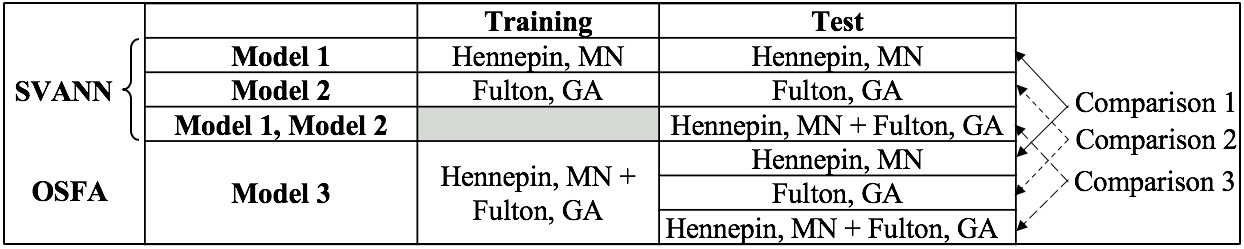}
   \caption{Assessment of spatial variability.}
  \label{fig:SV}
\end{figure}

Model 1 and Model 2 were SVANN-baed approach and Model 3 was OSFA-based approach. The evaluation on test data allow an "apples to apples" comparison of the three models. Figure \ref{fig:SV} shows the design to assess spatial variability. There were three sets of comparison: SVANN model is compared to OSFA using Hennepin County test data (OSFA\_HC); SVANN  Model 2 is compared to OSFA (OSFA\_FC) using Fulton County test data; Combined SVANN models 1 and 2 compared to OSFA Model 3 on the complete test dataset. Object characteristics are further used for finer level analysis. Framework parameters such as the number of iterations (I), image partitions (S), and IoU were assessed for the model performance. These results were used for tuning the model. The effect of transfer learning was also assessed on the model efficiency.

\textbf{Resources:}
\label{section:Resources}
The experiments were conducted using backpropagation algorithm using a python based Google Tensorflow variant of Darknet \cite{redmon2017yolo9000}. We used K40 GPU composed of 40 Haswell Xeon E5-2680 v3 nodes. Each node has 128 GB of RAM and 2 NVidia Tesla K40m GPUs. Each K40m GPU has 11 GB of RAM and 2880 CUDA cores.
\section{Experimental Results}
\label{section:RD}
This section presents our spatial variability assessment results and feature based interpretation.

\textbf{What is the effect of spatial variability? }
To assess the effect of spatial variability we considered three comparisons (Figure \ref{fig:SV}). Table \ref{Table:Comparison_1} shows the results. As shown, both SVANN Model 1 and SVANN Model 2 perform better than OSFA on all the measures (precision, recall, and F1-score) for all three comparisons. The results clearly demonstrate the effectiveness of SVANN over OSFA approach. The latter had learning samples from both regions, which may have resulted in a relatively larger generalization and low recognition of area specific objects. The effect can increase as spatial variation increases across the regions. This results in the dilution of regional differences that may otherwise be useful to identify the objects more accurately.
\label{subsection:variability}
\begin{table}[htp]
\small
\centering
\caption{Comparison results between SVANN and OSFA.}
\begin{tabular}{|p{1.15cm} p{1.1cm} p{1.3cm} p{1.10cm} p{0.75cm} p{1.15cm}|}
 \hline
 \textbf{Approach} & \textbf{Model} & \textbf{Test data} & \textbf{Precision} & \textbf{Recall} & \textbf{F1-score}\\
 \hline
  SVANN & Model 1 & Hennepin & 0.794 & 0.419 & 0.549\\
 OSFA & Model 3 & Hennepin & 0.713 & 0.341 & 0.461 \\
 \hline
 \hline
  SVANN & Model 2 & Fulton & 0.924 & 0.674 & 0.779\\
 OSFA & Model 3 & Fulton & 0.886 & 0.618 & 0.728 \\
 \hline
 \hline
 SVANN & Model 1, Model 2 & Hennepin + Fulton & 0.836 & 0.485 & 0.614\\
 OSFA & Model 3 & Hennepin + Fulton & 0.771 & 0.412 & 0.537 \\
 \hline
\end{tabular}
\label{Table:Comparison_1}
\end{table}
Due to limited training data for the application, we rely on external weights to build effective models. The results might have showed higher variability if we had not used external weights. In addition, spatial variability results that use minimal or no transfer learning can be used to further assess the effects and observe the variability in the results. 

Before these experiments our urban planning collaborators assume that urban gardens in Minnesota and Georgia are similar. They were surprised by our results that SVANN outperform OSFA and asked for detailed interpretation of our results. In this paragraph we summarize the findings from the detailed interpretation of results in context of urban garden detection.

\begin{figure}[htp]
  \centering
  \includegraphics[width=\linewidth]{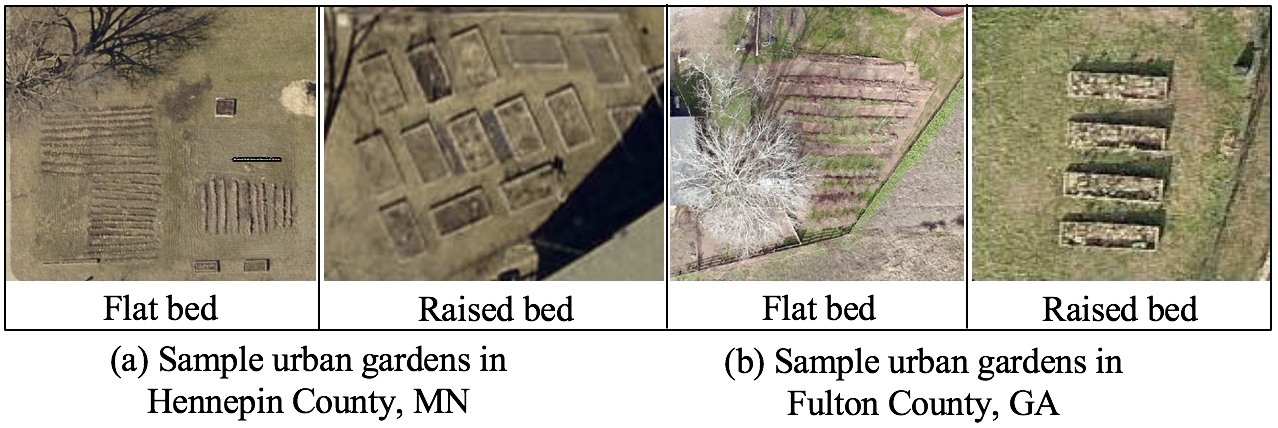}
  \caption{Spatial variability in the dataset. As shown, the backyard urban gardens in Fulton county, GA have greener surroundings compared to the backyard urban gardens in Hennepin county, MN. (Best in color).}
  \label{fig:examples}
\end{figure}

\textbf{Characteristic based interpretation:}
\label{subsection:interpretation}
As shown in Table \ref{Table:HC_data} and Table \ref{Table:FC_data}, Fulton County has a significantly higher proportion of raised beds to flat beds. This may suggest different gardening practices in the two regions. Further, this difference may have resulted in higher measure values for SVANN model 2 compared to SVANN model 1; because detection of raised beds is less challenging due to distinct boundaries. In terms of spatial variability, we found that gardens differed in their texture across the two regions. In particular, gardens in Fulton County, GA had a higher green cover as compared to Hennepin Couty, MN. The difference is highlighted in Figure \ref{fig:examples}, which depicts both raised and flat bed gardens from the two regions.

\section{Discussion}
\label{section:Related-Work}
Our approach (Section \ref{subsec:training} 2. distance weighted model-location dependent sampling for learning) is similar to Geographically Weighted Regression (GWR) \cite{fotheringham2003geographically} where regression coefficients and error are location dependent. However, GWR is a regression based technique that rely on manual features to calculate model weights. In contrast, we use multi-layer CNN \cite{lecun2015deep}, where initial layers perform feature engineering and later layers are responsible for prediction. 

The approach is also related to common practice in data mining where we first partition the data, and then develop prediction model separately for each partition. The partitions are formed in a high dimensional space which may mute geographic variability. In contrast, we use the partitions in low dimension geographic space in the proposed technique (Section \ref{subsec:training} 1a. Fixed partition based neighbors). Similar approach was followed in \cite{jiang2019spatial}, where a spatial ensemble framework was proposed that explicitly partitions input data in geographic space and use neighborhood effect to build models within each zone. Further, spatial variability has been discussed as a challenge to detect other geospatial objects such as trees \cite{xie2018timber, xie2019revolutionizing} and buildings \cite{xie2019locally, xie2020locally} using remote sensing datasets.

\section{Conclusion and Future Work}
\label{section:CFW}
In this work, we investigated a spatial-variability aware deep neural network (SVANN) approach where distinct deep neural network models are built for each geographic area. We also describe some of the training and prediction procedures for SVANN and list key points relevant to the approach. We chose high spatial resolution imagery for better object detection performance and built deep learning models using a state-of-the-art single stage object detection technique. We evaluated this approach using aerial imagery from two geographic areas for the task of mapping urban gardens. The experimental results show that SVANN provides better performance in terms of precision, recall, and F1-score to identify urban gardens. We also provide a case study that interprets spatial variability in terms of the relative frequency of urban garden characteristics. 

In the future, we plan to extend our evaluation of SVANN using other training and prediction approaches. We plan to generalize the proposed approach to model variability in high dimensional spaces. We also plan to evaluate SVANN with other types of neural networks, where we will evaluate the choice of models and network structure in terms of number of layers, neurons per layer, etc. Finally, we will evaluate the trade-off between spatial variability awareness and transfer learning.

\begin{acks}
This material is based upon work supported by the National Science Foundation under Grants No. 1029711, 1737633. We would like to thank Dr. Dana Boyer (Princeton University) and Dr. Anu Ramaswami (Princeton University) for providing the Hennepin county dataset and for useful guidance on Urban Agriculture. We would like to thank Minnesota Supercomputing Institute (MSI) for GPU resources. We would also like to thank Kim Koffolt and the spatial computing research group for their helpful comments and refinements.
\end{acks}

\bibliographystyle{ACM-Reference-Format}
\bibliography{deepspatial}


\begin{thebibliography}{38}


\ifx \showCODEN    \undefined \def \showCODEN     #1{\unskip}     \fi
\ifx \showDOI      \undefined \def \showDOI       #1{#1}\fi
\ifx \showISBNx    \undefined \def \showISBNx     #1{\unskip}     \fi
\ifx \showISBNxiii \undefined \def \showISBNxiii  #1{\unskip}     \fi
\ifx \showISSN     \undefined \def \showISSN      #1{\unskip}     \fi
\ifx \showLCCN     \undefined \def \showLCCN      #1{\unskip}     \fi
\ifx \shownote     \undefined \def \shownote      #1{#1}          \fi
\ifx \showarticletitle \undefined \def \showarticletitle #1{#1}   \fi
\ifx \showURL      \undefined \def \showURL       {\relax}        \fi
\providecommand\bibfield[2]{#2}
\providecommand\bibinfo[2]{#2}
\providecommand\natexlab[1]{#1}
\providecommand\showeprint[2][]{arXiv:#2}

\bibitem[\protect\citeauthoryear{Boyer, Kosse, Ambrose, Nixon, and
  Ramaswami}{Boyer et~al\mbox{.}}{2020}]%
        {Boyer2020}
\bibfield{author}{\bibinfo{person}{Dana Boyer}, \bibinfo{person}{Rachel Kosse},
  \bibinfo{person}{Graham Ambrose}, \bibinfo{person}{Peter Nixon}, {and}
  \bibinfo{person}{Anu Ramaswami}.} \bibinfo{year}{2020}\natexlab{}.
\newblock \showarticletitle{A hybrid transect \& remote sensing approach for
  mapping urban agriculture: informing food action plans \& metrics. --- under
  review}.
\newblock \bibinfo{journal}{\emph{Landscape and Urban Planning}}
  (\bibinfo{year}{2020}).
\newblock


\bibitem[\protect\citeauthoryear{Brown and McCarty}{Brown and McCarty}{2017}]%
        {brown2017remote}
\bibfield{author}{\bibinfo{person}{ME Brown} {and} \bibinfo{person}{JL
  McCarty}.} \bibinfo{year}{2017}\natexlab{}.
\newblock \showarticletitle{Is remote sensing useful for finding and monitoring
  urban farms?}
\newblock \bibinfo{journal}{\emph{Applied geography}}  \bibinfo{volume}{80}
  (\bibinfo{year}{2017}), \bibinfo{pages}{23--33}.
\newblock


\bibitem[\protect\citeauthoryear{Cybenko}{Cybenko}{1989}]%
        {cybenko1989approximations}
\bibfield{author}{\bibinfo{person}{George Cybenko}.}
  \bibinfo{year}{1989}\natexlab{}.
\newblock \showarticletitle{Approximations by superpositions of a sigmoidal
  function}.
\newblock \bibinfo{journal}{\emph{Mathematics of Control, Signals and Systems}}
   \bibinfo{volume}{2} (\bibinfo{year}{1989}), \bibinfo{pages}{183--192}.
\newblock


\bibitem[\protect\citeauthoryear{Erickson, Korfiatis, Akkus, and
  Kline}{Erickson et~al\mbox{.}}{2017}]%
        {erickson2017machine}
\bibfield{author}{\bibinfo{person}{Bradley~J Erickson},
  \bibinfo{person}{Panagiotis Korfiatis}, \bibinfo{person}{Zeynettin Akkus},
  {and} \bibinfo{person}{Timothy~L Kline}.} \bibinfo{year}{2017}\natexlab{}.
\newblock \showarticletitle{Machine learning for medical imaging}.
\newblock \bibinfo{journal}{\emph{Radiographics}} \bibinfo{volume}{37},
  \bibinfo{number}{2} (\bibinfo{year}{2017}), \bibinfo{pages}{505--515}.
\newblock


\bibitem[\protect\citeauthoryear{Fotheringham, Brunsdon, and
  Charlton}{Fotheringham et~al\mbox{.}}{2003}]%
        {fotheringham2003geographically}
\bibfield{author}{\bibinfo{person}{A~Stewart Fotheringham},
  \bibinfo{person}{Chris Brunsdon}, {and} \bibinfo{person}{Martin Charlton}.}
  \bibinfo{year}{2003}\natexlab{}.
\newblock \bibinfo{booktitle}{\emph{Geographically weighted regression: the
  analysis of spatially varying relationships}}.
\newblock \bibinfo{publisher}{John Wiley \& Sons}.
\newblock


\bibitem[\protect\citeauthoryear{Freund, Schapire, et~al\mbox{.}}{Freund
  et~al\mbox{.}}{1996}]%
        {freund1996experiments}
\bibfield{author}{\bibinfo{person}{Yoav Freund}, \bibinfo{person}{Robert~E
  Schapire}, {et~al\mbox{.}}} \bibinfo{year}{1996}\natexlab{}.
\newblock \showarticletitle{Experiments with a new boosting algorithm}. In
  \bibinfo{booktitle}{\emph{icml}}, Vol.~\bibinfo{volume}{96}. Citeseer,
  \bibinfo{pages}{148--156}.
\newblock


\bibitem[\protect\citeauthoryear{Gorelick, Hancher, Dixon, Ilyushchenko, Thau,
  and Moore}{Gorelick et~al\mbox{.}}{2017}]%
        {gorelick2017google}
\bibfield{author}{\bibinfo{person}{Noel Gorelick}, \bibinfo{person}{Matt
  Hancher}, \bibinfo{person}{Mike Dixon}, \bibinfo{person}{Simon Ilyushchenko},
  \bibinfo{person}{David Thau}, {and} \bibinfo{person}{Rebecca Moore}.}
  \bibinfo{year}{2017}\natexlab{}.
\newblock \showarticletitle{Google Earth Engine: Planetary-scale geospatial
  analysis for everyone}.
\newblock \bibinfo{journal}{\emph{Remote sensing of Environment}}
  \bibinfo{volume}{202} (\bibinfo{year}{2017}), \bibinfo{pages}{18--27}.
\newblock


\bibitem[\protect\citeauthoryear{Guo, Liu, Oerlemans, Lao, Wu, and Lew}{Guo
  et~al\mbox{.}}{2016}]%
        {guo2016deep}
\bibfield{author}{\bibinfo{person}{Yanming Guo}, \bibinfo{person}{Yu Liu},
  \bibinfo{person}{Ard Oerlemans}, \bibinfo{person}{Songyang Lao},
  \bibinfo{person}{Song Wu}, {and} \bibinfo{person}{Michael~S Lew}.}
  \bibinfo{year}{2016}\natexlab{}.
\newblock \showarticletitle{Deep learning for visual understanding: A review}.
\newblock \bibinfo{journal}{\emph{Neurocomputing}}  \bibinfo{volume}{187}
  (\bibinfo{year}{2016}), \bibinfo{pages}{27--48}.
\newblock


\bibitem[\protect\citeauthoryear{Gupta}{Gupta}{2020}]%
        {dataset_code}
\bibfield{author}{\bibinfo{person}{Jayant Gupta}.}
  \bibinfo{year}{2020}\natexlab{}.
\newblock \bibinfo{booktitle}{\emph{Urban Garden Dataset and SVANN code}}.
\newblock
\urldef\tempurl%
\url{https://tinyurl.com/yczutkjw}
\showURL{%
\tempurl}


\bibitem[\protect\citeauthoryear{Heindl, Nawaz, and Yuan}{Heindl
  et~al\mbox{.}}{2015}]%
        {heindl2015mapping}
\bibfield{author}{\bibinfo{person}{Andreas Heindl}, \bibinfo{person}{Sidra
  Nawaz}, {and} \bibinfo{person}{Yinyin Yuan}.}
  \bibinfo{year}{2015}\natexlab{}.
\newblock \showarticletitle{Mapping spatial heterogeneity in the tumor
  microenvironment: a new era for digital pathology}.
\newblock \bibinfo{journal}{\emph{Laboratory investigation}}
  \bibinfo{volume}{95}, \bibinfo{number}{4} (\bibinfo{year}{2015}),
  \bibinfo{pages}{377--384}.
\newblock


\bibitem[\protect\citeauthoryear{Jiang, Sainju, Li, Shekhar, and Knight}{Jiang
  et~al\mbox{.}}{2019}]%
        {jiang2019spatial}
\bibfield{author}{\bibinfo{person}{Zhe Jiang}, \bibinfo{person}{Arpan~Man
  Sainju}, \bibinfo{person}{Yan Li}, \bibinfo{person}{Shashi Shekhar}, {and}
  \bibinfo{person}{Joseph Knight}.} \bibinfo{year}{2019}\natexlab{}.
\newblock \showarticletitle{Spatial ensemble learning for heterogeneous
  geographic data with class ambiguity}.
\newblock \bibinfo{journal}{\emph{ACM Transactions on Intelligent Systems and
  Technology (TIST)}} \bibinfo{volume}{10}, \bibinfo{number}{4}
  (\bibinfo{year}{2019}), \bibinfo{pages}{1--25}.
\newblock


\bibitem[\protect\citeauthoryear{Krizhevsky, Sutskever, and Hinton}{Krizhevsky
  et~al\mbox{.}}{2012}]%
        {krizhevsky2012imagenet}
\bibfield{author}{\bibinfo{person}{Alex Krizhevsky}, \bibinfo{person}{Ilya
  Sutskever}, {and} \bibinfo{person}{Geoffrey~E Hinton}.}
  \bibinfo{year}{2012}\natexlab{}.
\newblock \showarticletitle{Imagenet classification with deep convolutional
  neural networks}. In \bibinfo{booktitle}{\emph{Advances in neural information
  processing systems}}. \bibinfo{pages}{1097--1105}.
\newblock


\bibitem[\protect\citeauthoryear{LeCun, Bengio, and Hinton}{LeCun
  et~al\mbox{.}}{2015}]%
        {lecun2015deep}
\bibfield{author}{\bibinfo{person}{Y LeCun}, \bibinfo{person}{Y Bengio}, {and}
  \bibinfo{person}{G Hinton}.} \bibinfo{year}{2015}\natexlab{}.
\newblock \showarticletitle{Deep learning}.
\newblock \bibinfo{journal}{\emph{nature}} \bibinfo{volume}{521},
  \bibinfo{number}{7553} (\bibinfo{year}{2015}), \bibinfo{pages}{436}.
\newblock


\bibitem[\protect\citeauthoryear{Leitner, Glasner, and Kounadi}{Leitner
  et~al\mbox{.}}{2018}]%
        {leitner2018laws}
\bibfield{author}{\bibinfo{person}{Michael Leitner}, \bibinfo{person}{Philip
  Glasner}, {and} \bibinfo{person}{Ourania Kounadi}.}
  \bibinfo{year}{2018}\natexlab{}.
\newblock \showarticletitle{Laws of geography}.
\newblock In \bibinfo{booktitle}{\emph{Oxford Research Encyclopedia of
  Criminology and Criminal Justice}}.
\newblock


\bibitem[\protect\citeauthoryear{Lin, Maire, Belongie, Hays, Perona, Ramanan,
  Doll{\'a}r, and Zitnick}{Lin et~al\mbox{.}}{2014}]%
        {lin2014microsoft}
\bibfield{author}{\bibinfo{person}{Tsung-Yi Lin}, \bibinfo{person}{Michael
  Maire}, \bibinfo{person}{Serge Belongie}, \bibinfo{person}{James Hays},
  \bibinfo{person}{Pietro Perona}, \bibinfo{person}{Deva Ramanan},
  \bibinfo{person}{Piotr Doll{\'a}r}, {and} \bibinfo{person}{C~Lawrence
  Zitnick}.} \bibinfo{year}{2014}\natexlab{}.
\newblock \showarticletitle{Microsoft coco: Common objects in context}. In
  \bibinfo{booktitle}{\emph{European conference on computer vision}}. Springer,
  \bibinfo{pages}{740--755}.
\newblock


\bibitem[\protect\citeauthoryear{Manning, Raghavan, and Sch{\"u}tze}{Manning
  et~al\mbox{.}}{2008}]%
        {manning2008introduction}
\bibfield{author}{\bibinfo{person}{Christopher~D Manning},
  \bibinfo{person}{Prabhakar Raghavan}, {and} \bibinfo{person}{Hinrich
  Sch{\"u}tze}.} \bibinfo{year}{2008}\natexlab{}.
\newblock \bibinfo{booktitle}{\emph{Introduction to information retrieval}}.
\newblock \bibinfo{publisher}{Cambridge university press}.
\newblock


\bibitem[\protect\citeauthoryear{Mathieu, Freeman, and Aryal}{Mathieu
  et~al\mbox{.}}{2007}]%
        {mathieu2007mapping}
\bibfield{author}{\bibinfo{person}{Renaud Mathieu}, \bibinfo{person}{Claire
  Freeman}, {and} \bibinfo{person}{Jagannath Aryal}.}
  \bibinfo{year}{2007}\natexlab{}.
\newblock \showarticletitle{Mapping private gardens in urban areas using
  object-oriented techniques and very high-resolution satellite imagery}.
\newblock \bibinfo{journal}{\emph{Landscape and Urban Planning}}
  \bibinfo{volume}{81}, \bibinfo{number}{3} (\bibinfo{year}{2007}),
  \bibinfo{pages}{179--192}.
\newblock


\bibitem[\protect\citeauthoryear{Miotto, Wang, Wang, Jiang, and Dudley}{Miotto
  et~al\mbox{.}}{2018}]%
        {miotto2018deep}
\bibfield{author}{\bibinfo{person}{Riccardo Miotto}, \bibinfo{person}{Fei
  Wang}, \bibinfo{person}{Shuang Wang}, \bibinfo{person}{Xiaoqian Jiang}, {and}
  \bibinfo{person}{Joel~T Dudley}.} \bibinfo{year}{2018}\natexlab{}.
\newblock \showarticletitle{Deep learning for healthcare: review, opportunities
  and challenges}.
\newblock \bibinfo{journal}{\emph{Briefings in bioinformatics}}
  \bibinfo{volume}{19}, \bibinfo{number}{6} (\bibinfo{year}{2018}),
  \bibinfo{pages}{1236--1246}.
\newblock


\bibitem[\protect\citeauthoryear{of~Agriculture}{of~Agriculture}{2019}]%
        {DATA_NAIP}
\bibfield{author}{\bibinfo{person}{US~Department of Agriculture}.}
  \bibinfo{year}{2019}\natexlab{}.
\newblock \bibinfo{title}{Geospatial data gateway}.
\newblock
\newblock
\urldef\tempurl%
\url{https://datagateway.nrcs.usda.gov}
\showURL{%
Retrieved February 27, 2020 from \tempurl}


\bibitem[\protect\citeauthoryear{of~Agriculture}{of~Agriculture}{2012}]%
        {USDA_hardiness}
\bibfield{author}{\bibinfo{person}{United States~Department of Agriculture}.}
  \bibinfo{year}{2012}\natexlab{}.
\newblock \bibinfo{booktitle}{\emph{USDA Plant Hardiness Zone Map}}.
\newblock USDA.
\newblock
\urldef\tempurl%
\url{https://planthardiness.ars.usda.gov/PHZMWeb/}
\showURL{%
Retrieved 2019-12-23 from \tempurl}


\bibitem[\protect\citeauthoryear{Qiu}{Qiu}{2019}]%
        {Bbox_label}
\bibfield{author}{\bibinfo{person}{Shi Qiu}.} \bibinfo{year}{2019}\natexlab{}.
\newblock \bibinfo{booktitle}{\emph{Bbox label tool}}.
\newblock
\urldef\tempurl%
\url{https://github.com/puzzledqs}
\showURL{%
Retrieved 2019-12-23 from \tempurl}


\bibitem[\protect\citeauthoryear{Redmon and Farhadi}{Redmon and
  Farhadi}{2017}]%
        {redmon2017yolo9000}
\bibfield{author}{\bibinfo{person}{Joseph Redmon} {and} \bibinfo{person}{Ali
  Farhadi}.} \bibinfo{year}{2017}\natexlab{}.
\newblock \showarticletitle{YOLO9000: better, faster, stronger}. In
  \bibinfo{booktitle}{\emph{Proceedings of the IEEE conference on computer
  vision and pattern recognition}}. \bibinfo{pages}{7263--7271}.
\newblock


\bibitem[\protect\citeauthoryear{Russell and Norvig}{Russell and
  Norvig}{2002}]%
        {russell2002artificial}
\bibfield{author}{\bibinfo{person}{Stuart Russell} {and} \bibinfo{person}{Peter
  Norvig}.} \bibinfo{year}{2002}\natexlab{}.
\newblock \showarticletitle{Artificial intelligence: a modern approach}.
\newblock  (\bibinfo{year}{2002}).
\newblock


\bibitem[\protect\citeauthoryear{Shekhar, Gunturi, Evans, and Yang}{Shekhar
  et~al\mbox{.}}{2012}]%
        {shekhar2012spatial}
\bibfield{author}{\bibinfo{person}{Shashi Shekhar}, \bibinfo{person}{Viswanath
  Gunturi}, \bibinfo{person}{Michael~R Evans}, {and} \bibinfo{person}{KwangSoo
  Yang}.} \bibinfo{year}{2012}\natexlab{}.
\newblock \showarticletitle{Spatial big-data challenges intersecting mobility
  and cloud computing}. In \bibinfo{booktitle}{\emph{Proceedings of the
  Eleventh ACM International Workshop on Data Engineering for Wireless and
  Mobile Access}}. \bibinfo{pages}{1--6}.
\newblock


\bibitem[\protect\citeauthoryear{Silvertown}{Silvertown}{2009}]%
        {silvertown2009new}
\bibfield{author}{\bibinfo{person}{Jonathan Silvertown}.}
  \bibinfo{year}{2009}\natexlab{}.
\newblock \showarticletitle{A new dawn for citizen science}.
\newblock \bibinfo{journal}{\emph{Trends in ecology \& evolution}}
  \bibinfo{volume}{24}, \bibinfo{number}{9} (\bibinfo{year}{2009}),
  \bibinfo{pages}{467--471}.
\newblock


\bibitem[\protect\citeauthoryear{Spinhirne}{Spinhirne}{1993}]%
        {spinhirne1993micro}
\bibfield{author}{\bibinfo{person}{James~D Spinhirne}.}
  \bibinfo{year}{1993}\natexlab{}.
\newblock \showarticletitle{Micro pulse lidar}.
\newblock \bibinfo{journal}{\emph{IEEE Transactions on Geoscience and Remote
  Sensing}} \bibinfo{volume}{31}, \bibinfo{number}{1} (\bibinfo{year}{1993}),
  \bibinfo{pages}{48--55}.
\newblock


\bibitem[\protect\citeauthoryear{Systems}{Systems}{2019}]%
        {DATA_FC}
\bibfield{author}{\bibinfo{person}{Fulton County Geographic~Information
  Systems}.} \bibinfo{year}{2019}\natexlab{}.
\newblock \bibinfo{title}{The Aerial Imagery Download Tool}.
\newblock
\newblock
\urldef\tempurl%
\url{https://gis.fultoncountyga.gov/apps/AerialDownloadMapViewer/}
\showURL{%
Retrieved February 05, 2020 from \tempurl}


\bibitem[\protect\citeauthoryear{Systems}{Systems}{2015}]%
        {DATA_HC}
\bibfield{author}{\bibinfo{person}{Hennepin County Geographic~Information
  Systems}.} \bibinfo{year}{2015}\natexlab{}.
\newblock \bibinfo{title}{Hennepin County Aerial Imagery}.
\newblock
\newblock
\urldef\tempurl%
\url{https://gis.fultoncountyga.gov/apps/AerialDownloadMapViewer/}
\showURL{%
Retrieved December 27, 2020 from \tempurl}


\bibitem[\protect\citeauthoryear{Szegedy, Liu, Jia, Sermanet, Reed, Anguelov,
  Erhan, Vanhoucke, and Rabinovich}{Szegedy et~al\mbox{.}}{2015}]%
        {szegedy2015going}
\bibfield{author}{\bibinfo{person}{Christian Szegedy}, \bibinfo{person}{Wei
  Liu}, \bibinfo{person}{Yangqing Jia}, \bibinfo{person}{Pierre Sermanet},
  \bibinfo{person}{Scott Reed}, \bibinfo{person}{Dragomir Anguelov},
  \bibinfo{person}{Dumitru Erhan}, \bibinfo{person}{Vincent Vanhoucke}, {and}
  \bibinfo{person}{Andrew Rabinovich}.} \bibinfo{year}{2015}\natexlab{}.
\newblock \showarticletitle{Going deeper with convolutions}. In
  \bibinfo{booktitle}{\emph{Proceedings of the IEEE conference on computer
  vision and pattern recognition}}. \bibinfo{pages}{1--9}.
\newblock


\bibitem[\protect\citeauthoryear{Turner and Chapin}{Turner and Chapin}{2005}]%
        {turner2005causes}
\bibfield{author}{\bibinfo{person}{Monica~G Turner} {and}
  \bibinfo{person}{F~Stuart Chapin}.} \bibinfo{year}{2005}\natexlab{}.
\newblock \showarticletitle{Causes and consequences of spatial heterogeneity in
  ecosystem function}.
\newblock In \bibinfo{booktitle}{\emph{Ecosystem function in heterogeneous
  landscapes}}. \bibinfo{publisher}{Springer}, \bibinfo{pages}{9--30}.
\newblock


\bibitem[\protect\citeauthoryear{Wagner}{Wagner}{1982}]%
        {wagner1982simpson}
\bibfield{author}{\bibinfo{person}{Clifford~H Wagner}.}
  \bibinfo{year}{1982}\natexlab{}.
\newblock \showarticletitle{Simpson's paradox in real life}.
\newblock \bibinfo{journal}{\emph{The American Statistician}}
  \bibinfo{volume}{36}, \bibinfo{number}{1} (\bibinfo{year}{1982}),
  \bibinfo{pages}{46--48}.
\newblock


\bibitem[\protect\citeauthoryear{Woodcock, Allen, Anderson, Belward,
  Bindschadler, Cohen, Gao, Goward, Helder, Helmer, et~al\mbox{.}}{Woodcock
  et~al\mbox{.}}{2008}]%
        {woodcock2008free}
\bibfield{author}{\bibinfo{person}{Curtis~E Woodcock}, \bibinfo{person}{Richard
  Allen}, \bibinfo{person}{Martha Anderson}, \bibinfo{person}{Alan Belward},
  \bibinfo{person}{Robert Bindschadler}, \bibinfo{person}{Warren Cohen},
  \bibinfo{person}{Feng Gao}, \bibinfo{person}{Samuel~N Goward},
  \bibinfo{person}{Dennis Helder}, \bibinfo{person}{Eileen Helmer},
  {et~al\mbox{.}}} \bibinfo{year}{2008}\natexlab{}.
\newblock \showarticletitle{Free access to Landsat imagery}.
\newblock \bibinfo{journal}{\emph{Science}} \bibinfo{volume}{320},
  \bibinfo{number}{5879} (\bibinfo{year}{2008}), \bibinfo{pages}{1011--1011}.
\newblock


\bibitem[\protect\citeauthoryear{Xie, Bao, Shekhar, and Knight}{Xie
  et~al\mbox{.}}{2018a}]%
        {xie2018timber}
\bibfield{author}{\bibinfo{person}{Yiqun Xie}, \bibinfo{person}{Han Bao},
  \bibinfo{person}{Shashi Shekhar}, {and} \bibinfo{person}{Joseph Knight}.}
  \bibinfo{year}{2018}\natexlab{a}.
\newblock \showarticletitle{A TIMBER Framework for Mining Urban Tree
  Inventories Using Remote Sensing Datasets}. In \bibinfo{booktitle}{\emph{2018
  IEEE International Conference on Data Mining (ICDM)}}. IEEE,
  \bibinfo{pages}{1344--1349}.
\newblock


\bibitem[\protect\citeauthoryear{Xie, Cai, Bhojwani, Shekhar, and Knight}{Xie
  et~al\mbox{.}}{2019a}]%
        {xie2019locally}
\bibfield{author}{\bibinfo{person}{Y Xie}, \bibinfo{person}{J Cai},
  \bibinfo{person}{R Bhojwani}, \bibinfo{person}{S Shekhar}, {and}
  \bibinfo{person}{J Knight}.} \bibinfo{year}{2019}\natexlab{a}.
\newblock \showarticletitle{A locally-constrained YOLO framework for detecting
  small and densely-distributed building footprints}.
\newblock \bibinfo{journal}{\emph{International Journal of Geographical
  Information Science}} (\bibinfo{year}{2019}), \bibinfo{pages}{1--25}.
\newblock


\bibitem[\protect\citeauthoryear{Xie, Cai, Bhojwani, Shekhar, and Knight}{Xie
  et~al\mbox{.}}{2020}]%
        {xie2020locally}
\bibfield{author}{\bibinfo{person}{Yiqun Xie}, \bibinfo{person}{Jiannan Cai},
  \bibinfo{person}{Rahul Bhojwani}, \bibinfo{person}{Shashi Shekhar}, {and}
  \bibinfo{person}{Joseph Knight}.} \bibinfo{year}{2020}\natexlab{}.
\newblock \showarticletitle{A locally-constrained yolo framework for detecting
  small and densely-distributed building footprints}.
\newblock \bibinfo{journal}{\emph{International Journal of Geographical
  Information Science}} \bibinfo{volume}{34}, \bibinfo{number}{4}
  (\bibinfo{year}{2020}), \bibinfo{pages}{777--801}.
\newblock


\bibitem[\protect\citeauthoryear{Xie, Gupta, Li, and Shekhar}{Xie
  et~al\mbox{.}}{2018b}]%
        {xie2018transforming}
\bibfield{author}{\bibinfo{person}{Yiqun Xie}, \bibinfo{person}{Jayant Gupta},
  \bibinfo{person}{Yan Li}, {and} \bibinfo{person}{Shashi Shekhar}.}
  \bibinfo{year}{2018}\natexlab{b}.
\newblock \showarticletitle{Transforming Smart Cities with Spatial Computing}.
  In \bibinfo{booktitle}{\emph{2018 IEEE International Smart Cities Conference
  (ISC2)}}. IEEE, \bibinfo{pages}{1--9}.
\newblock


\bibitem[\protect\citeauthoryear{Xie, Shekhar, Feiock, and Knight}{Xie
  et~al\mbox{.}}{2019b}]%
        {xie2019revolutionizing}
\bibfield{author}{\bibinfo{person}{Yiqun Xie}, \bibinfo{person}{Shashi
  Shekhar}, \bibinfo{person}{Richard Feiock}, {and} \bibinfo{person}{Joseph
  Knight}.} \bibinfo{year}{2019}\natexlab{b}.
\newblock \showarticletitle{Revolutionizing tree management via intelligent
  spatial techniques}. In \bibinfo{booktitle}{\emph{Proceedings of the 27th ACM
  SIGSPATIAL International Conference on Advances in Geographic Information
  Systems}}. \bibinfo{pages}{71--74}.
\newblock


\bibitem[\protect\citeauthoryear{Zhu, Tuia, Mou, Xia, Zhang, Xu, and
  Fraundorfer}{Zhu et~al\mbox{.}}{2017}]%
        {zhu2017deep}
\bibfield{author}{\bibinfo{person}{Xiao~Xiang Zhu}, \bibinfo{person}{Devis
  Tuia}, \bibinfo{person}{Lichao Mou}, \bibinfo{person}{Gui-Song Xia},
  \bibinfo{person}{Liangpei Zhang}, \bibinfo{person}{Feng Xu}, {and}
  \bibinfo{person}{Friedrich Fraundorfer}.} \bibinfo{year}{2017}\natexlab{}.
\newblock \showarticletitle{Deep learning in remote sensing: A comprehensive
  review and list of resources}.
\newblock \bibinfo{journal}{\emph{IEEE Geoscience and Remote Sensing Magazine}}
  \bibinfo{volume}{5}, \bibinfo{number}{4} (\bibinfo{year}{2017}),
  \bibinfo{pages}{8--36}.
\newblock


\end{thebibliography}

\appendix
\section{Aerial imagery}
\label{Appendix:Aerial}
\begin{figure}[htp]
  \centering
    \includegraphics[width=\linewidth]{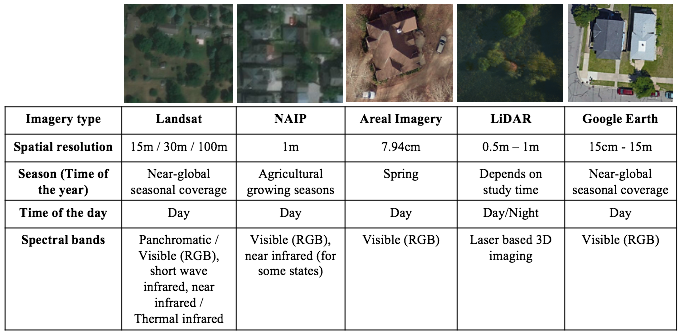}
  \caption{Imagery types}
  \label{fig:Imagery_Types}
\end{figure}

Unlike the related work ~\cite{zhu2017deep, mathieu2007mapping}, which is based on satellite imagery with one-meter order resolution, we use higher resolution (7.5cm) aerial imagery taken at the start of spring season. Figure \ref{fig:Imagery_Types} shows the types of imagery that we considered, their resolution, season, time of the day, spectral bands and a visual example. It shows 5 types of imagery namely, Landsat \cite{woodcock2008free}, National Agriculture Imagery Program (NAIP) \cite{DATA_NAIP}, areal imagery \cite{DATA_FC, DATA_HC}, LIDAR point cloud \cite{spinhirne1993micro}, and Google Earth imagery \cite{gorelick2017google}. As shown, areal imagery taken in the beginning of the spring season is better to detect urban objects due to lower occlusion from leaves and snow cover. In addition, higher resolution results in sharper images that allow better detection of distinctive features. For this work, we have not used LIDAR data for training, as it may add sensor based variability in addition to the geographic variability. Further, satellite based imagery and the available Google Earth were not used due to seasonal cover and relatively low resolution respectively.

\section{You Only Look Once (YOLO) for object detection}
\label{Appendix:YOLO}
SVANN uses YOLO deep learning framework to train multiple models. Here, we briefly describe its architecture, object detection procedure, and a subset of relevant parameters.

\textbf{Architecture \cite{redmon2017yolo9000}:} The framework has 24 convolution layers for feature engineering and selection, and two fully connected layers for prediction. The framework uses $3\times3$ filters to extract features and 1$\times$1 filters to reduce output channels. The prediction layer has two fully-connected layers that perform linear regression on the final two layers to make the boundary box predictions to detect objects. The first layer flattens the 3-dimensional vector output from the convolution layer to a single dimension 4096 vector. The final layer converts the 1D vector to a 3D vector with the detected values. 

\begin{figure}[htp]
  \centering
   \includegraphics[width=\linewidth]{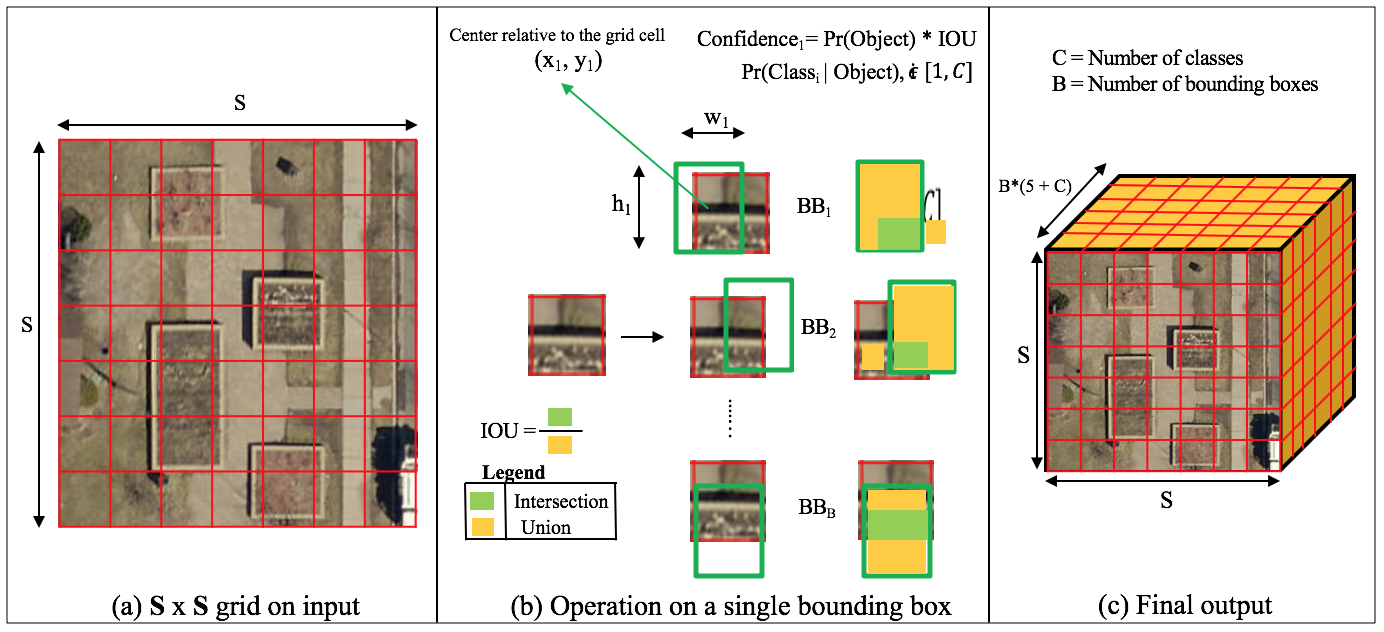}
   \caption{Object detection in YOLO (Best in color)}
  \label{fig:Image_Detection}
\end{figure}

The input to the framework is a $448\times448$ dimension image, which is a product of a prime number ($7$) and multiple of 2. This allows the reduction of the dimensions by 2 across the convolution and pooling layers. The reduction is also affected by stride, that is, the step size to move the convolution matrix. The dimension reduces by half whenever pooling (e.g., Maxpool) and convolution layer with stride 2 is used. The channels increase with the dimension in the convolution filters.\\
\textbf{Object detection in YOLO:} Figure~\ref{fig:Image_Detection} shows the object detection sequence in YOLO~\cite{redmon2017yolo9000} using one of our example images. The first step is to break the image into an $S$x$S$ grid. Then, $B$ bounding boxes are predicted for each grid cell. B is a parameter provided to the framework using the value suggested in the YOLO paper. Each bounding box is represented by $4$ values ($x$, $y$, $w$, $h$) and a confidence score. All the $5$ values ($x$, $y$, $w$, $h$, and confidence score) are predicted by the fully connected layers. The $(x, y)$ coordinates represent the center of the bounding box, $w$ represents the width, and $h$ represents the height of the box relative to the complete image. The confidence score reflects the accuracy of the detected object. It is the product of the probability of the object and the Intersection over Union (IoU), where IoU is the ratio of the ground-truth and prediction intersection over their union (Figure \ref{fig:Image_Detection}(b)). The box with the highest IoU is selected. 
\\
\textbf{Framework parameters} affect the training time of the model and the number of predictions from the prediction layer. Each iteration of the gradient descent updates all the weights in the layers, which is time consuming. Further, after a certain number of iterations there is no significant change in precision and recall values. Besides iteration, the number of partitions for detection and IoU values affects the number of predictions. A lower number of partitions may result in limited detection in high density partitions. Further, higher IoU values may result in limited detection of small objects, whereas lower IoU values can increase the number of False Positives in the detection process.

\section{Dataset development}
\label{Appendix:data_development}

\begin{table}[htp]
\small
\centering
\caption{Comparison results between SVANN and OSFA.}
\begin{tabular}{|p{1.35cm} p{1.85cm} p{1.25cm} p{1.10cm} p{0.75cm} p{1.15cm}|}
 \hline
 \textbf{Approach} & \textbf{Training Area} & \textbf{Test Area} & \textbf{Precision} & \textbf{Recall} & \textbf{F1-score}\\
 \hline
  \textcolor{ForestGreen}{SVANN} & Hennepin & Hennepin & \textcolor{ForestGreen}{0.794} & \textcolor{ForestGreen}{0.419} & \textcolor{ForestGreen}{0.549}\\
 \textcolor{red}{OSFA} & All & Hennepin & \textcolor{red}{0.713} & \textcolor{red}{0.341} & \textcolor{red}{0.461} \\
 \hline
 \hline
  \textcolor{ForestGreen}{SVANN} & Fulton & Fulton & \textcolor{ForestGreen}{0.924} & \textcolor{ForestGreen}{0.674} & \textcolor{ForestGreen}{0.779}\\
 \textcolor{red}{OSFA} & All & Fulton & \textcolor{red}{0.886} & \textcolor{red}{0.618} & \textcolor{red}{0.728} \\
 \hline
 \hline
 \textcolor{ForestGreen}{SVANN (Two zones)} & All (trained \phantom{test} by zone) & Hennepin + Fulton & \textcolor{ForestGreen}{0.836} & \textcolor{ForestGreen}{0.485} & \textcolor{ForestGreen}{0.614}\\
 \textcolor{red}{OSFA} & All & Hennepin + Fulton & \textcolor{red}{0.771} & \textcolor{red}{0.412} & \textcolor{red}{0.537} \\
 \hline
\end{tabular}
\label{Table:Comparison_1}
\end{table}
\subsection{Conversion to high spatial resolution aerial imagery}
\label{Appendix:Conversion}
\begin{figure}[htp]
  \centering
    \includegraphics[width=0.80\linewidth]{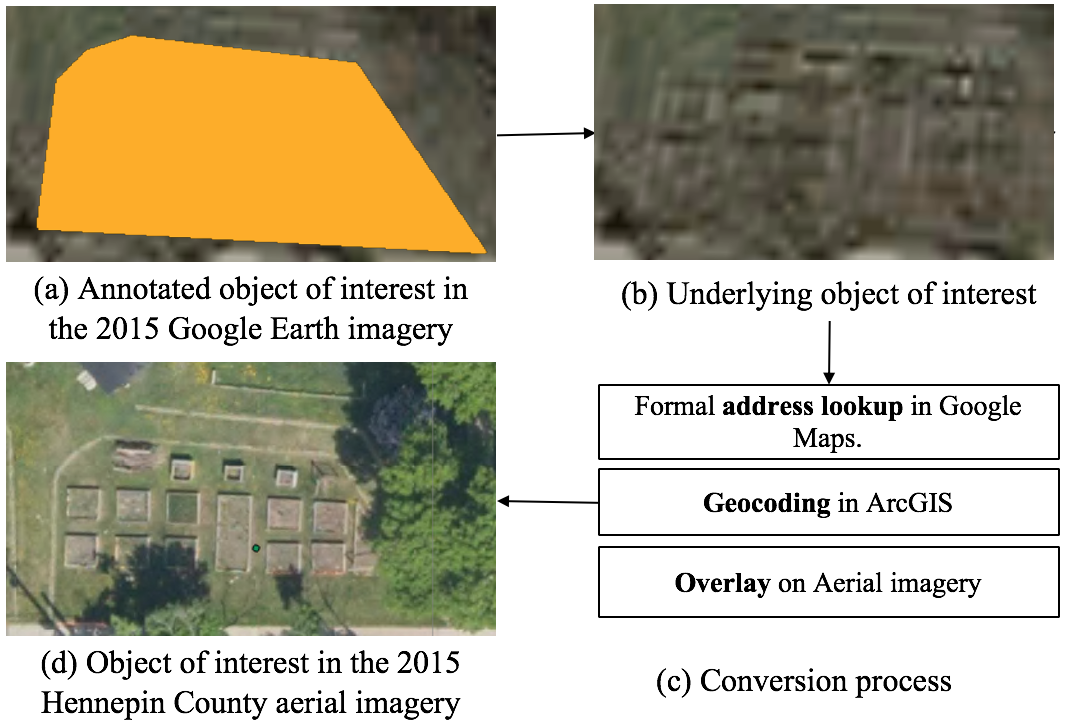}
   \caption{Conversion of existing annotations.}
  \label{fig:Conversion}
\end{figure}

The key challenge was inconsistency in the naming, such as, use of informal names (e.g., 3437 Garden) that could not be mapped to the latitude and longitude. Thus, we visually looked up the objects in Google Earth and through the nearest road intersection identified the formal location using Google Maps (3437 S 15th Ave Minneapolis, MN 55407). The locations were Geocoded in ArcGIS and overlayed on the high spatial resolution aerial imagery to extract improved object of interest samples. Figure \ref{fig:Conversion} shows the annotated input, underlying object of interest, conversion process, and the object of interest in the high spatial resolution imagery. 
\subsection{Manual annotations from high spatial resolution aerial imagery}
\label{Appendix:Annotation}
\begin{figure}[htp]
  \centering
   \includegraphics[width=0.95\linewidth]{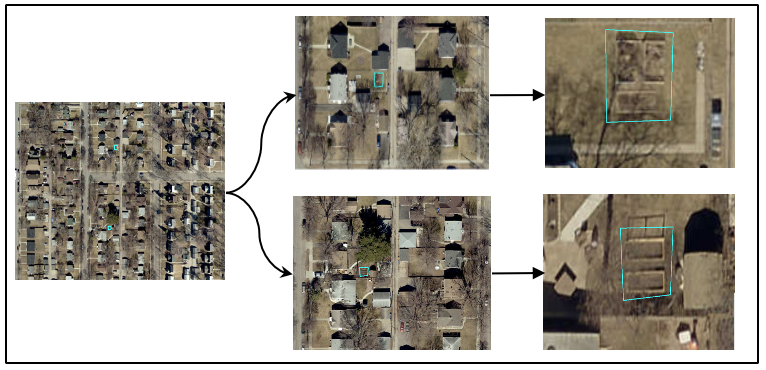}
   \caption{Image annotation sequence.}
  \label{fig:AnnotationSequence}
\end{figure}

Figure \ref{fig:AnnotationSequence} shows the image annotation sequence where the zoom level increases from left to right. The area of the object was annotated in a 4-step process: First, geo-tagging the object; Second, creating a rectangular buffer; Third, clipping the object; and Fourth, annotating the object using a bounding box. Annotations were done using BBox-Label-Tool \cite{Bbox_label}, a python based annotation tool.

\end{document}